**An iterative energy-based multimodal transformer for joint retrieval of wheat soil moisture, leaf area index, and plant height from Sentinel-1 and Sentinel-2 time series**

Shubham Kumar Singh[1*], Peilei Fan[1], Suraj A. Yadav[2], Rajendra Prasad[3], Prashant K Srivastava[4]

[1]Department of Urban and Environmental Policy and Program, Tufts University, 503 Boston Ave., Medford, MA 02155
[2]Electrical and Computer Engineering Department, Mississippi State University, Mississippi, USA, 39762
[3]Department of Physics, Indian Institute of Technology (BHU), Varanasi, India 221005
[4]Institute of Environment and Sustainable Development, Banaras Hindu University, Varanasi, India. 221005

**Abstract**

Field-scale retrieval of surface soil moisture (SM), leaf area index (LAI), and plant height (PH) is essential for precision agriculture, yet it remains an ill-posed inverse problem. Concurrent variations in soil moisture and canopy density generate substantial ambiguities in radar backscatter and spectral responses, which reduces the effectiveness of traditional feed-forward regression models in heterogeneous smallholder cropping systems. This study presents the Iterative Energy-Based Transformer (iEBT) for the joint retrieval of coupled soil-canopy states from Sentinel-1 C-band SAR and Sentinel-2 multispectral time series. Instead of direct regression, iEBT embeds multi-modal predictors within a shared sequence, produces an initial state estimate, and iteratively updates the target [SM, LAI, PH] vector through normalized gradient descent to minimize a learned scalar compatibility energy function. Using 700 quality-controlled field measurements from Varanasi, India, iEBT achieved the highest learned-model performance on the random test split, with a four-seed mean $R^2$ of 0.854 ± 0.012 ($R^2_{SM} = 0.841, R^2_{LAI} = 0.905, R^2_{PH} = 0.821$). WCM and PROSAIL were retained as physically interpretable SAR and optical reference models for comparison. Modality ablations confirmed that Sentinel-1 drives SM retrieval while Sentinel-2 dominates LAI, whereas PH relies on combined structural-phenological signatures. Crucially, the model's terminal energy functions as an uncalibrated post-retrieval quality diagnostic; screening the 10% highest-energy samples markedly reduced target-level root-mean-square errors. While leave-one-campaign-out validation highlights persistent cross-season domain shift challenges due to localized management variations, compatibility-guided multi-modal fusion offers a structured, self-diagnostic path toward reliable biophysical parameter estimation.

Code availability: The implementation of iEBT and experimental scripts is available at: GitHub

## 1. Introduction

Field-scale monitoring of crop biophysical status is central to precision agriculture because management decisions increasingly require spatially explicit information on water availability, canopy development, and structural crop growth (Pan et al., 2019; Zhuo et al., 2019). In wheat systems, surface soil moisture (SM), leaf area index (LAI), and plant height (PH) reveal information about the coupled soil-crop system (Pan et al., 2019; Wang et al., 2024; Papadavid & Toulios, 2017). SM determines the availability of near-surface soil water and the dielectric properties of the soil (Corbari et al., 2020; Santi & Paloscia, 2019). LAI is a measure of the density of the canopy and its ability to intercept radiation, whereas PH indicates cumulative crop structural development and was measured directly at each growth stage. (Bouras et al., 2020; Lin et al., 2022; Liu et al., 2023; Frappart et al., 2020). Because irrigation, management practices, and phenological development vary among fields, joint retrieval of SM, LAI, and PH provides a more complete description of wheat condition than estimating each variable independently (ex. wheat cultivation practices within the Indo-Gangetic Plain). Despite the need for field data for validation and calibration, field data cannot provide the necessary spatial coverage over the diverse agricultural lands. Satellite remote sensing offers a practical route for monitoring wheat water status, canopy growth, and structural development at the field scale (Ahmad et al., 2022; Pierdicca et al., 2022; Bateni et al., 2013).

Satellite remote sensing observations, such as Sentinel-1 synthetic aperture radar (SAR) and Sentinel-2 multispectral imagery, provide complementary sensitivity to these crop and soil properties (Singh et al., 2023; Nduku et al., 2024). Sentinel-2 reflectance in the visible and near-infrared bands, together with vegetation indices, is strongly related to canopy greenness, chlorophyll absorption, vegetation density, and LAI (Bahrami et al., 2022). However, optical retrieval is constrained by cloud cover, haze, aerosol effects, illumination-viewing geometry, and saturation of vegetation indices under dense canopy conditions (Veloso et al., 2017). Sentinel-1 C-band SAR provides an important complementary observation because it is sensitive to soil dielectric properties, surface roughness, vegetation water content, and canopy structure, and it can acquire observations independent of solar illumination and with reduced sensitivity to atmospheric conditions (Alliès et al., 2021; Ghosh et al., 2022). In agricultural fields, VV backscatter is generally more responsive to soil-surface and soil-canopy interaction scattering, whereas VH backscatter is more closely associated with vegetation volume scattering and crop structural development. These sensor-specific responses make SAR-optical fusion attractive for wheat biophysical retrieval (Singh et al., 2024; Veloso et al., 2017).

Despite this complementarity, retrieving SM, LAI, and PH from satellite observations remains an ill-posed inverse problem (Liu et al., 2019; Steele-Dunne et al., 2017). Similar SAR or optical responses can arise from different combinations of soil moisture, canopy density, surface roughness, vegetation water content, plant height, and phenological stage (Li & Wang, 2018; Zribi et al., 2019; Chauhan et al., 2018). In SAR observations, the soil contribution to backscatter becomes increasingly attenuated as the wheat canopy develops, reducing the separability of soil-moisture and vegetation effects (Steele-Dunne et al., 2017; Bousbih et al., 2017). In optical observations, canopy reflectance and vegetation indices can lose sensitivity when the canopy becomes dense. Temporal mismatch between field measurements and satellite acquisitions further increases uncertainty, especially during rapid phenological transitions or after irrigation events (Hajj et al., 2016; Hosseini et al., 2021). These ambiguities mean that a field-scale retrieval model must not only learn nonlinear relationships between satellite predictors and crop variables, but also handle multimodal evidence that may be noisy, incomplete, temporally mismatched, or internally inconsistent (Zhang et al., 2021; Yang et al., 2021).

Physical and semi-empirical models provide an interpretable foundation for remote-sensing retrieval. Optical radiative transfer models such as PROSAIL link canopy reflectance to leaf optical properties, canopy architecture, and illumination-viewing geometry, and are widely used for estimating vegetation variables such as LAI (Tao et al., 2019). Microwave models such as the Water Cloud Model represent vegetation as an attenuating layer and decompose radar backscatter into vegetation and soil contributions (Bouchat et al., 2022; Gou et al., 2022; Li & Wang, 2018; Zhang et al., 2018), providing a physically meaningful framework for vegetation-covered soil-moisture retrieval. These approaches are valuable because their assumptions are explicit and their parameters have physical interpretation (Zribi et al., 2019). However, their operational use over heterogeneous wheat fields is limited by simplified canopy representations (Vermunt et al., 2022; Park et al., 2019), site-specific calibration requirements (Hajj et al., 2016), sensitivity to roughness and vegetation descriptors (Das & Pandey, 2024; Li & Wang, 2018), and difficulty in retrieving multiple coupled soil-crop variables simultaneously (Zribi et al., 2019). PROSAIL is mainly suited to optical canopy variables, whereas WCM relies on SAR scattering assumptions that can become unstable when soil and vegetation contributions are strongly coupled ( Bousbih et al., 2017; Steele-Dunne et al., 2017).

Machine-learning and deep-learning approaches address some of these limitations by learning nonlinear relationships directly from field-satellite observations (Meng et al., 2018; Han et al., 2020; Liu et al., 2021). Random forests, neural networks, and transformer-based models can integrate SAR, optical, and temporal predictors without requiring a fully specified physical forward model (Ayehu et al., 2020; Hosseini et al., 2021). Multimodal learning is especially relevant because Sentinel-1 and Sentinel-2 encode different components of the crop system: radar observations provide moisture and structural sensitivity, while optical reflectance and vegetation indices provide strong canopy information (Li & Wang, 2018; Zhang et al., 2021). However, most data-driven retrieval models remain direct regressors. They map the fused predictor vector to SM, LAI, and PH in a single forward pass, but they do not explicitly evaluate whether the predicted crop state is compatible with the combined microwave, optical, and temporal evidence (Hajj et al., 2016; Yang et al., 2021). This limitation is important for operational precision agriculture, where residual cloud contamination, local roughness, irrigation events, acquisition mismatch, or cross-season distribution shift can produce plausible but unreliable predictions (Beale et al., 2019).

Energy-based learning offers a structured alternative for this retrieval problem. Rather than learning only a direct mapping from observations to target variables, an energy-based model learns a scalar compatibility function over observation-state pairs, assigning low energy to plausible input-output configurations and high energy to incompatible ones (LeCun et al., 2006; Belanger & McCallum, 2016). Prediction can then be formulated as learned inversion: the observations are fixed, and the candidate [SM, LAI, PH] state is adjusted to minimize the learned energy, following the broader EBM view of inference as optimization over output variables (Tu & Gimpel, 2019). This idea is conceptually related to physical inversion, but the compatibility surface is learned from data rather than specified entirely through an explicit radiative transfer or scattering model. Recent Energy-Based Transformer work further motivates this design by embedding iterative energy minimization within transformer-based architectures for learned refinement and reasoning (Gladstone et al., 2025).

In this study, we develop an iterative energy-based transformer for joint retrieval of wheat SM, LAI, and PH from Sentinel-1 SAR, Sentinel-2 optical, and temporal features. The model first encodes SAR, optical, and temporal predictors using modality-aware representations, generates an initial crop-state proposal, and then refines the candidate [SM, LAI, PH] vector by minimizing a learned compatibility energy conditioned on the multimodal observations. Rather than treating SAR-optical fusion as simple feature concatenation, the proposed framework evaluates whether a candidate soil-canopy state is consistent with the combined microwave, optical, and temporal evidence. The iterative refinement process also exposes the transition from the initial proposal to the final retrieved state, while the terminal energy provides an uncalibrated diagnostic of retrieval compatibility.

## 2. Materials and methods

### 2.1. Study area and field campaigns

Field observations were collected from wheat fields in Varanasi district, Uttar Pradesh, India, located in the central Indo-Gangetic Plain. Wheat is widely cultivated during the rabi season, typically from late November or early December to April. The region includes irrigated agricultural fields and alluvial soils with variable water-holding capacity. Its relatively flat terrain reduces topographic effects in SAR observations, enabling field-scale comparisons between radar backscatter, optical reflectance, and in situ crop measurements. Although the field campaigns were conducted within one district, the site represents irrigated, small-field wheat systems of the Indo-Gangetic Plain, where heterogeneous management, irrigation timing, and rapid phenological transitions create challenging conditions for SAR-optical crop-state retrieval.

The data used in this paper were obtained from fields and satellites during three wheat-growing seasons: 2019-2020, 2023, and 2024 (Table 1). These campaigns aimed to capture wheat phenological differences, irrigation, canopy density and crop structural development. The observations for the 2023 and 2024 campaigns were from January to April, while for 2019-2020, they were limited to the early vegetative period of late December to early January. Sampling in this period gave pre-canopy baseline data with which to evaluate the retrieval framework in various growth stages and seasonal trajectories. The final data set consisted of 700 field-satellite sampling events after strict quality control. The locations of the campaigns and context at the district-level are shown in Fig. 1 along with the sampling distributions of locations at the annual level.

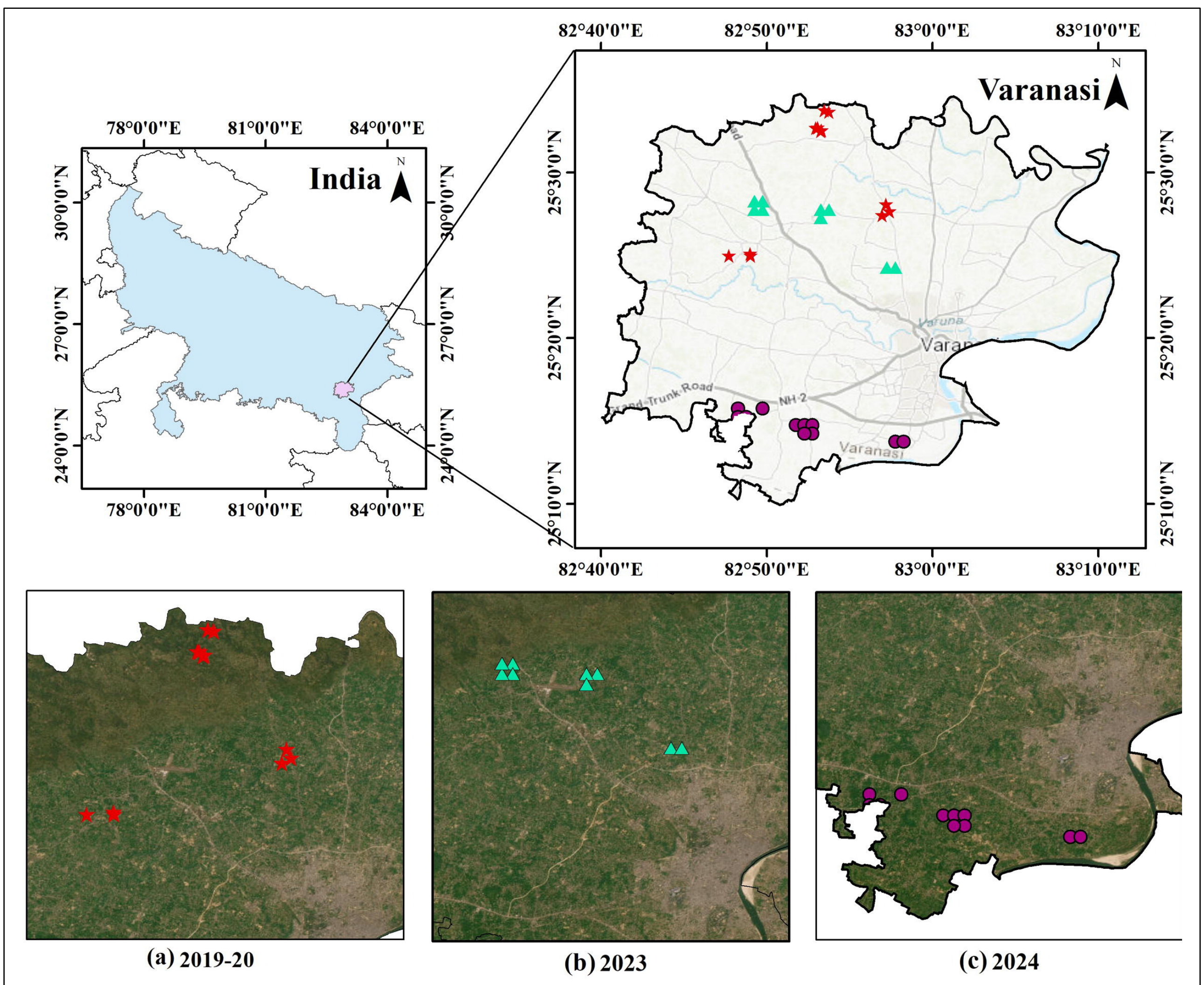

Fig. 1. The study area and field campaign locations. The top panel depicts the geographical location of Varanasi district in northern India and the sampling sites where field observations were collected, including markers for each campaign (upper right). The bottom three panels (a), (b) and (c) display sample locations from 2019-2020 wheat campaign, 2023 wheat campaign, and 2024 wheat campaign, respectively.

**Table 1: Field Measurement campaign and dates**

| Campaign | Site location label | Field sampling dates | Sentinel-1 acquisition dates | Sentinel-2 acquisition date | No. of strict-QC observations |
|---|---|---|---|---|---|
| **2019-20** | Wheat field | 2019-12-22; 2020-01-11; 2020-02-11 | 2019-12-22; 2020-01-11; 2020-02-11 | 2019-12-22; 2020-01-11; 2020-02-10 | 184 |
| **2023** | BHU Wheat field | 2023-01-23; 2023-02-15; 2023-02-16; 2023-02-27; 2023-02-28; 2023-03-10; 2023-03-13; 2023-03-23; 2023-03-24 | 2023-01-22; 2023-02-15; 2023-02-27; 2023-03-11; 2023-03-23 | 2023-01-25; 2023-02-14; 2023-03-01; 2023-03-11; 2023-03-21; 2023-03-26 | 165 |
| | Pindra wheat field | 2023-01-23; 2023-02-03; 2023-02-16; 2023-02-28; 2023-03-13; 2023-03-23 | 2023-01-22; 2023-02-03; 2023-02-15; 2023-02-27; 2023-03-11; 2023-03-23 | 2023-01-25; 2023-02-04; 2023-02-14; 2023-03-01; 2023-03-11; 2023-03-21 | 160 |
| **2024** | BHU wheat | 2024-02-19; 2024-03-28; 2024-04-09 | 2024-02-19; 2024-03-29; 2024-04-07 | 2024-02-19; 2024-03-30; 2024-04-09 | 35 |
| | Pindra wheat | 2024-02-08; 2024-02-19; 2024-03-05; 2024-03-14; 2024-03-28; 2024-04-09 | 2024-02-07; 2024-02-19; 2024-03-05; 2024-03-14; 2024-03-29; 2024-04-07 | 2024-02-09; 2024-02-19; 2024-03-05; 2024-03-15; 2024-03-30; 2024-04-09 | 156 |

## 2.2 Field measurements

SM, LAI, and PH were measured three times in each sampling area. These variables are chosen as they represent complementary parts to the soil-crop system and are generally predicted to affect both microwave scattering and optical reflectance. The soil dielectric permittivity and radar backscatter are influenced by SM, the canopy density and optical absorption-scattering processes by LAI, and the crop structural development by PH. Surface SM was measured as volumetric water content, i.e., the ratio of the volume of water to the total volume of soil, in units of $m^3/m^3$ using Hydra-Go probe device. Measurements were taken from the upper soil layer at 0-5 cm. For each sampling plot, 3 replicate measurements were collected and averaged to reduce local-scale soil heterogeneity. LAI, the ratio of green leaf area to the ground surface area, was measured in units of $m^2/ m^2$ using LAI-2200 (LI-COR, Inc.). Measurements were collected under sunny days, where possible, to reduce directional illumination effects. For each field plot, readings were collected three times and averaged. PH was measured manually in centimeters from the soil surface to the top of the wheat canopy using a ruler.

Quality control was applied to remove observations with missing field variables, invalid satellite features, cloud-contaminated Sentinel-2 observations, large temporal mismatch, or inconsistent measurements. The final strict-QC dataset contained 700 observations. Summary statistics for the target variables are reported in Table 2.

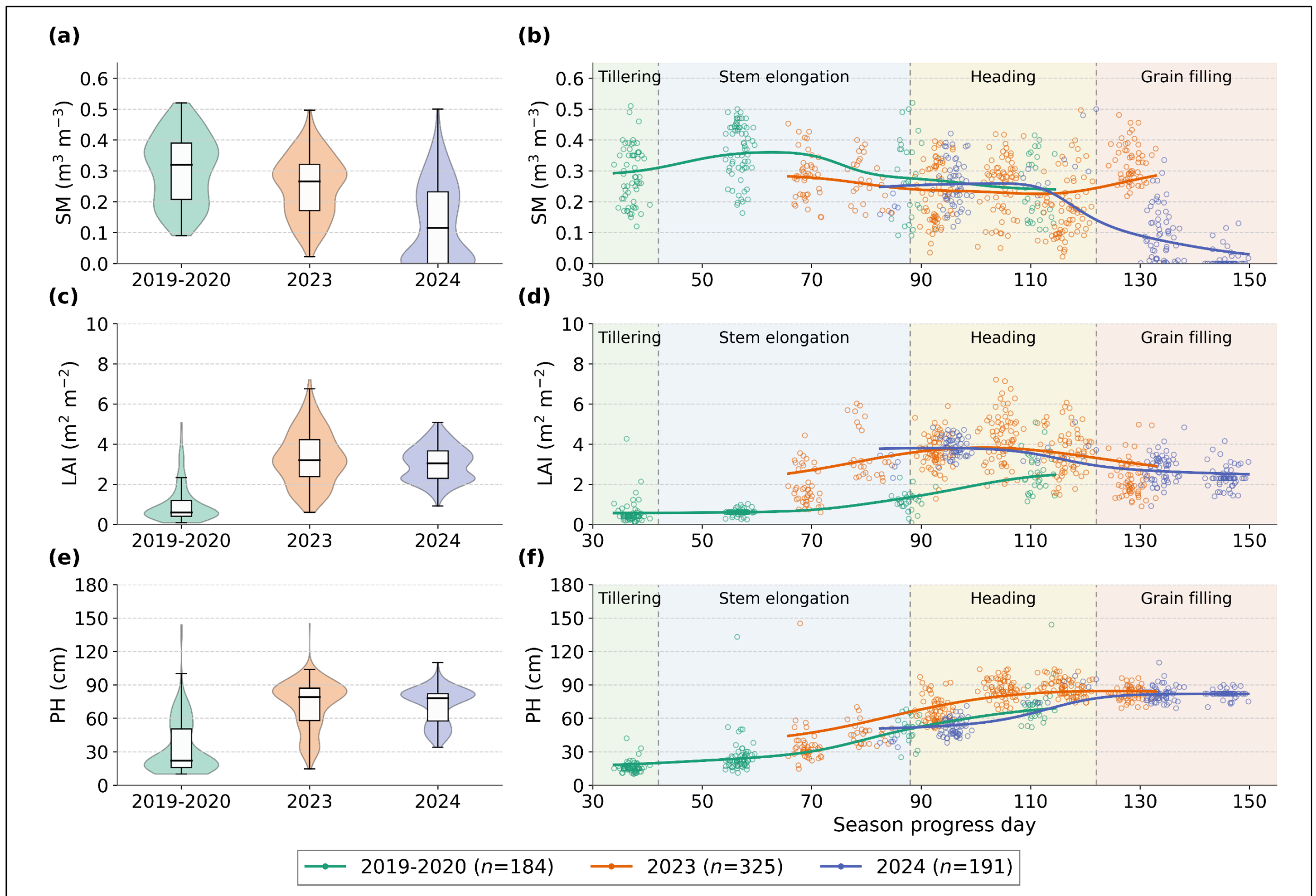

Fig. 2. Field-measured target distributions and phenological trajectories across wheat campaigns. Panels (a), (c), and (e) show campaign-level distributions of surface soil moisture (SM), leaf area index (LAI), and plant height (PH), respectively. Panels (b), (d), and (f) show the same variables as a function of season-progress day. Colors distinguish the 2019-2020, 2023, and 2024 campaigns; points represent field-satellite observations; trend lines summarize campaign-specific seasonal trajectories; shaded backgrounds indicate approximate wheat growth stages.

**Table 2. Target-variable statistics for campaigns and random-split subsets.**

| **Subset** | **n** | **SM mean ± std. ($m^3/m^3$)** | **LAI mean ± std. ($m^2/m^2$)** | **PH mean ± std. (cm)** |
|---|---|---|---|---|
| 2019-2020 campaign | 184 | 0.306 ± 0.111 | 0.82 ± 1.05 | 30.9 ± 26.3 |
| 2023 campaign | 325 | 0.250 ± 0.099 | 3.31 ± 1.35 | 72.2 ± 21.5 |
| 2024 campaign | 191 | 0.133 ± 0.129 | 2.71 ± 1.34 | 51.3 ± 33.4 |
| All observations | 700 | 0.233 ± 0.128 | 2.49 ± 1.64 | 55.6 ± 31.5 |
| Training set | 490 | 0.232 ± 0.128 | 2.55 ± 1.62 | 56.5 ± 31.4 |
| Validation set | 105 | 0.234 ± 0.135 | 2.42 ± 1.75 | 54.5 ± 33.1 |
| Test set | 105 | 0.236 ± 0.126 | 2.26 ± 1.61 | 52.7 ± 30.7 |

**Notes**: SM = soil moisture; LAI = leaf area index; PH = plant height. Training, validation, and test statistics correspond to the 70/15/15 random split used for the main benchmark.

## 2.3 Sentinel-1 and Sentinel-2 preprocessing

Sentinel-1 C-band SAR observations were used to provide microwave sensitivity to soil moisture, surface roughness, vegetation water content, and canopy structure. Interferometric Wide Swath Ground Range Detected products were used in dual-polarization mode, including vertical transmit-vertical receive (VV) and vertical transmit-horizontal receive (VH) backscatter. The Sentinel-1 preprocessing chain included SNAPPY, ESA SNAP, Google Earth Engine, Python/rasterio pipeline. Three SAR predictors were retained: VV backscatter, VH backscatter, and the VV/VH polarization ratio. VV was used to represent co-polarized surface and soil-canopy interaction scattering, VH was used to represent cross-polarized vegetation volume scattering, and VV/VH was included as a compact descriptor of changing canopy structure.

Sentinel-2 optical observations were used to capture canopy spectral response during wheat development. Level-2A surface reflectance products were used to provide atmospherically corrected reflectance values. Four spectral bands were retained: blue (B2), green (B3), red (B4), and near-infrared (B8). These bands were selected because they capture chlorophyll absorption, canopy greenness, and near-infrared scattering from leaf and canopy structure. The SWIR band B11 was used only to compute the vegetation water index and was resampled to the common analysis resolution when required. Cloud and cloud-shadow contamination were removed using the Sen2Cor scene classification mask. Reflectance values were extracted from the point buffer and the plot.

Three vegetation indices were computed as follows (Equations 1, 2, and 3)

$$NDVI = \frac{B8 - B4}{B8 + B4} \quad (1)$$

$$EVI = 2.5\frac{B8 - B4}{B8 + 6B4 - 7.5B2 + 1} \quad (2)$$

$$NDMI = \frac{B08 - B11}{B08 + B11} \quad (3)$$

where B2, B4, B8, B11 are the spectral bands of Blue (B2), Red (B4), Near-Infrared (B8), and short Waved Infrared (SWIR) (B11) of Sentinel-2.

## 2.4. Temporal matching and predictor construction

Each field observation was paired with the closest valid Sentinel-1 and Sentinel-2 observations using a nearest-date temporal matching strategy. Field-satellite pairs were retained only when both Sentinel-1 and Sentinel-2 observations were available within a maximum temporal window of ±5 days. The acquisition date gaps were calculated separately for Sentinel-1 and Sentinel-2 as Equation (4) and Equation (5):

$$\Delta t_{S1} = t_{S1} - t_f \quad (4)$$
$$\Delta t_{S2} = t_{S2} - t_f \quad (5)$$

where $t_f$ is the field sampling date, $t_{S1}$ is the Sentinel-1 acquisition date, and $t_{S2}$ is the Sentinel-2 acquisition date. Signed acquisition gaps were retained as predictors to represent the temporal mismatch between field and satellite observations.

Because wheat growth spans the December-January calendar-year boundary, crop timing was represented using a season-relative variable rather than calendar day of year. Season-progress day $d_s$ was defined as the number of days since 15 November of each wheat season. Cyclic seasonal encodings were then computed as Equation (6) and Equation (7):

$$s_d = \sin\left(\frac{2\pi d_s}{180}\right) \quad (6)$$
$$c_d = \cos\left(\frac{2\pi d_s}{180}\right) \quad (7)$$

The final predictor vector contained 15 variables: three SAR variables, seven optical variables, and five temporal variables (Table 3). Campaign identifiers and absolute sampling dates were not used as input predictors in the main retrieval models; campaign labels were used only for stratified interpretation and leave-one-campaign-out validation.

**Table 3. Predictor variables used for multimodal retrieval.**

| Group | Variable | Symbol / Name | Source | Unit / Scale | Physical/Modeling Role |
|---|---|---|---|---|---|
| SAR | VV backscatter | VV | Sentinel-1 | dB $\sigma^0$ | Co-polarized surface and canopy scattering |
| | VH backscatter | VH | | dB $\sigma^0$ | Cross-polarized vegetation volume scattering |
| | Polarization ratio | VV/VH | | Unitless | Proxy for canopy structure and scattering balance |
| Optical | Blue reflectance | B2 | Sentinel-2 | Surface reflectance | Atmospheric/aerosol-sensitive visible band |
| | Green reflectance | B3 | | | Canopy greenness and chlorophyll-related response |
| | Red reflectance | B4 | | | Chlorophyll absorption |
| | Near-infrared reflectance | B8 | | | Leaf and canopy scattering |
| Optical index | Normalized difference vegetation index | NDVI | Sentinel-2 derived | Unitless | Canopy greenness and LAI-related signal |
| | Enhanced vegetation index | EVI | Sentinel-2 derived | | Canopy density with reduced soil/background effect |
| | Normalized difference water index | NDMI | Sentinel-2 derived | | Canopy/vegetation water-sensitive NIR-SWIR index |
| Temporal | Sentinel-1 date gap | $\Delta t_{S1}$ | Derived | Days | Signed Temporal mismatch between field and SAR observation |
| | Sentinel-2 date gap | $\Delta t_{S2}$ | Derived | | Signed Temporal mismatch between field and optical observation |
| | Season progress day | $d_{\text{season}}$ | Derived | | Rabi- Season Phenological progression |
| | Seasonal sin Encoding | sin(DSP) | Derived | unitless | Cyclic representation of within season timing |
| | Seasonal cosine Encoding | cos(DSP) | Derived | unitless | Cyclic representation of within season timing |

### 2.5. Retrieval formulation and iterative energy-based transformer

The retrieval task was formulated as supervised multi-output regression from multimodal satellite and temporal predictors to field-measured wheat biophysical variables. For the sampling event $i$, the target vector was defined in Equation (8) as

$$\mathbf{y}_i = \begin{bmatrix} SM_i \\ LAI_i \\ PH_i \end{bmatrix} \quad (8)$$

The input vector was decomposed into SAR, optical, and temporal components according to Equation (9):

$$\mathbf{x}_i = \{\mathbf{x}_{s,i}, \mathbf{x}_{o,i}, \mathbf{x}_{t,i}\} \quad (9)$$

where

$$\mathbf{x}_{s,i} \in \mathbb{R}^3, \mathbf{x}_{o,i} \in \mathbb{R}^7, \mathbf{x}_{t,i} \in \mathbb{R}^5.$$

All input features and target variables were standardized using parameters fitted only on the training set. The same transformations were applied to validation, test, modality-ablation, and leave-one-campaign-out folds.

#### 2.5.1. Multimodal encoding and proposal prediction

The proposed iEBT model uses modality-specific encoders followed by transformer-based fusion (Fig. 3). SAR, optical, and temporal inputs as in Equation (10) were mapped separately to 64-dimensional tokens:

$$\mathbf{z}_s = f_s(\mathbf{x}_s) \in \mathbb{R}^{64}, \mathbf{z}_o = f_o(\mathbf{x}_o) \in \mathbb{R}^{64}, \mathbf{z}_t = f_t(\mathbf{x}_t) \in \mathbb{R}^{64} \quad (10)$$

Each encoder used a linear projection, layer normalization, GELU activation, and dropout. The modality tokens were stacked into a shared multimodal sequence as Equation (11):

$$\mathbf{Z} = \text{stack}(\mathbf{z}_s, \mathbf{z}_o, \mathbf{z}_t) \in \mathbb{R}^{3\times 64} \quad (11)$$

A transformer fusion module learned cross-modal interactions:

$$\mathbf{H} = T_\phi(\mathbf{Z})$$

where $T_\phi$ denotes the transformer encoder. The proposal head produced an initial crop-state estimate defined as in Equation (12)

$$\hat{\mathbf{y}}^{(0)} = g_p(\mathbf{H}) \quad (12)$$

where $g_p$is a multilayer perceptron. This proposal served as the initialization for energy-based refinement.

#### 2.5.2. Energy function and iterative refinement

The energy function assigns a scalar compatibility score defined in Equation (13) to an observation-target pair :

$$E_\theta(\mathbf{Z}, \mathbf{y}) \in \mathbb{R} \quad (13)$$

Low energy indicates that the candidate [SM, LAI, PH] state is compatible with the SAR, optical, and temporal observation context. High energy indicates a less compatible candidate state.

At the refinement step $k$, the current target estimate was embedded using a target encoder as in Equation (14):

$$\mathbf{u}^{(k)} = h_y\left(\hat{\mathbf{y}}^{(k)}\right) \in \mathbb{R}^{64} \quad (14)$$

The target token was appended to the observation tokens defined in Equation (15) and passed through the energy transformer:

$$E^{(k)} = g_E\left(T_E\left([\mathbf{z}_s, \mathbf{z}_o, \mathbf{z}_t, \mathbf{u}^{(k)}]\right)\right) \quad (15)$$

where $T_E$ is the energy-fusion transformer and $g_E$ is the scalar energy head. The candidate target vector was updated by normalized gradient descent on the learned energy surface according to Equation (16):

$$\hat{\mathbf{y}}^{(k+1)} = \text{clip}\left[\hat{\mathbf{y}}^{(k)} - \alpha_k \frac{\nabla_{\mathbf{y}} E_\theta(\mathbf{Z}, \hat{\mathbf{y}}^{(k)})}{\| \nabla_{\mathbf{y}} E_\theta(\mathbf{Z}, \hat{\mathbf{y}}^{(k)}) \|_2 + \epsilon}\right] \quad (16)$$

Here, $\alpha_k$ is the refinement step size, $\epsilon$ is a numerical stability constant, and clip$(\cdot)$ constrains candidate states within the physically plausible standardized target range. Candidate states were clipped to ±4 standard deviations in the standardized target space. The final prediction after $K$ refinement steps were calculated as Equation (17)

$$\hat{\mathbf{y}}^* = \hat{\mathbf{y}}^{(K)} \quad (17)$$

The final reported iEBT inference used $K = 8$ refinement steps. The step-depth analysis is reported in the Supplementary, where $K = 8$ produced the lowest terminal energy and the highest average $R^2$ for this dataset.

#### 2.5.3. Training objective

The iEBT model was trained using proposal supervision, energy ranking, and final refined-target supervision. Equation (18) defines the proposal loss as

$$\mathcal{L}_{prop} = \frac{1}{B} \sum_{i=1}^{B} \| \hat{\mathbf{y}}_i^{(0)} - \mathbf{y}_i \|_2^2 \quad (18)$$

where $B$ is the batch size.

The energy-ranking loss encouraged true observation target pairs to have lower energy than corrupted pairs. For each observation, a negative target vector $\mathbf{y}_i^-$ , which was generated using the same-campaign target shuffling or proposal-centered perturbation. Equations (19) and (20) define the positive and negative energies as

$$E_i^+ = E_\theta(\mathbf{Z}_i, \mathbf{y}_i) \quad (19)$$
$$E_i^- = E_\theta(\mathbf{Z}_i, \mathbf{y}_i^-) \quad (20)$$

To ensure true observation-target pairs were assigned lower energy than corrupted pairs, the framework minimized the margin-ranking loss formulated in Equation (21)

$$\mathcal{L}_{rank} = \frac{1}{B} \sum_{i=1}^{B} \max(0, m + E_i^+ - E_i^-) \quad (21)$$

where $m$ is the energy margin. Same-campaign negative sampling was used to prevent the energy model from learning only trivial campaign-level separation, while proposal-centered perturbations created harder negatives near plausible crop states.

The final refined prediction as defined in Equation (22) was supervised using an uncertainty-weighted multi-target regression loss

$$\mathcal{L}_{final} = \sum_{j=1}^{3} \left[ \exp(-s_j) \frac{1}{B} \sum_{i=1}^{B} \left( \hat{y}_{ij}^{*} - y_{ij} \right)^2 + s_j \right] \quad (22)$$

where $j$ indexes SM, LAI, and PH, and $s_j$ is a learnable log-variance parameter. These learned weights balance target contributions in standardized space and are not interpreted as calibrated physical prediction intervals.

The complete system parameters were optimized jointly by minimizing the combined total objective function presented in Equation (23)

$$\mathcal{L} = \mathcal{L}_{final} + \lambda_{prop} \mathcal{L}_{prop} + \lambda_{rank} \mathcal{L}_{rank} \quad (23)$$

Model parameters were optimized using AdamW with learning-rate scheduling, weight decay, dropout, gradient clipping, and early stopping based on validation loss. Full optimizer settings, hyperparameters, seed list, and software versions are provided in Supplementary Table S3.

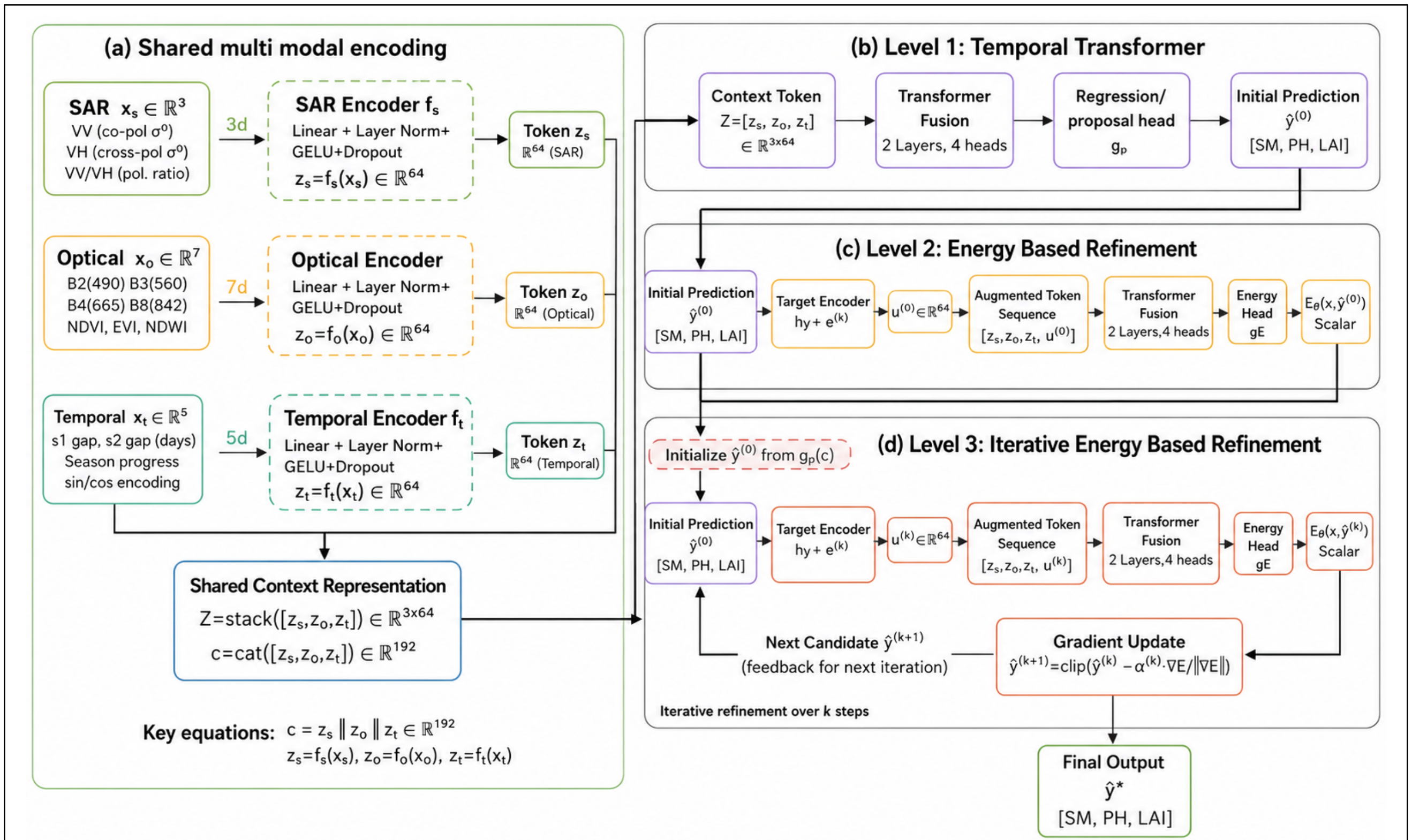

**Fig. 3. Multimodal iterative energy-based transformer architecture.** Panel (a) shows modality-specific encoding of Sentinel-1 SAR, Sentinel-2 optical, and temporal predictors into 64-dimensional tokens. Panel (b) shows the direct temporal-transformer proposal model. Panel (c) shows the non-iterative energy-based transformer baseline. Panel (d) shows the proposed iterative energy-based transformer, where an initial [SM, LAI, PH] proposal is refined by gradient-based minimization of a learned compatibility energy. Panel (e) summarizes feature fusion and the implementation of the prediction head. The final iEBT output is the refined

## 2.6. Baseline retrieval models

The proposed iEBT was compared with learned baselines and physically motivated reference models. These models are the Water Cloud Model (WCM) and PROSAIL. The WCM was used as a semi-empirical reference for Sentinel-1 SAR defined in Equation (24). For polarization $p \in \{VV, VH\}$, WCM was written as

$$\sigma_p^0 = A_p V \cos\theta \left(1 - \exp\left[\frac{2B_p V}{\cos\theta}\right]\right) + \left(C_p + D_p SM\right) \exp\left[\frac{2B_p V}{\cos\theta}\right] \quad (24)$$

$\sigma_{p,veg}^0$ and $\sigma_{p,soil}^0$ defined in Equations (25) and (26), respectively, are vegetation and surface backscattering responses and defined as

$$\sigma_{p,veg}^0 = A_p V \cos\theta \left(1 - \exp\left[\frac{2B_p V}{\cos\theta}\right]\right) \quad (25)$$

$$\sigma^0_{p,soil} = \left(C_p + D_p SM\right) \exp\left[\frac{2B_p V}{cos\,\theta}\right] \qquad (26)$$

where $\sigma_p^0$ is the radar backscatter coefficient, $V$ is the vegetation descriptor, $SM$ is volumetric surface soil moisture, $\theta$ is the incidence angle, and $A_p$, $B_p$, $C_p$, and $D_p$ are fitted polarization-specific parameters. The model parameters are described in Supplementary Section S6. The incidence angle was fixed at 38° for the WCM analysis. Parameters were fitted independently for VV and VH using bounded trust-region least-squares optimization. Inversion was performed using a one-dimensional grid search with 200 nodes. WCM was used as a physically interpretable SAR-only reference, not as a split-identical multimodal competitor.

The SNAP Biophysical Processor, which is based on neural-network inversion of PROSAIL simulations, was used as an optical physical reference for LAI retrieval from Sentinel-2. It was not used for SM or PH because the implemented product does not directly retrieve those variables. PROSAIL was therefore interpreted as an LAI-oriented optical reference rather than as a full multi-target competitor. Full WCM parameter bounds, fitted coefficients, inversion settings, and PROSAIL processing details are reported in Supplementary Section S6.

### 2.7. Experimental design and evaluation

The main model comparison used a random 70/15/15 split into training, validation, and test sets. This produced 490 training observations, 105 validation observations, and 105 held-out test observations. The random split was used to evaluate within-distribution retrieval performance.

Leave-one-campaign-out validation was used to evaluate cross-campaign transferability. In each fold, models were trained on two campaigns and tested on the held-out campaign. This experiment tested whether retrieval models could generalize across years with different phenological distributions, irrigation conditions, crop-growth trajectories, and sensor-response patterns.

A modality-ablation experiment was used to quantify the contribution of Sentinel-1 SAR and Sentinel-2 optical information. Three input configurations were evaluated: SAR-only, optical-only, and full multimodal input. Temporal variables were retained in all ablation settings because timing information is essential for wheat phenological interpretation.

For neural models, all benchmark experiments were repeated across four random seeds. Quantitative performance tables report the mean ± standard deviation across seeds. Observed-versus-estimated scatter plots and refinement diagnostics are shown for one representative prediction set and are intended for qualitative interpretation rather than quantitative model ranking.

Model performance was evaluated using $R^2$ and RMSE defined in Equations (27) and (28), respectively. For target $j$, these metrics were computed as

$$R_j^2 = 1 - \frac{\sum_i (y_{ij} - \hat{y}_{ij})^2}{\sum_i (y_{ij} - \bar{y}_j)^2} \qquad (27)$$

$$RMSE_j = \sqrt{\frac{1}{n}\sum_i (y_{ij} - \hat{y}_{ij})^2} \quad (28)$$

Average $R^2$ was computed as the arithmetic mean across SM, LAI, and PH according to Equation (29) as

$$\bar{R^2} = \frac{1}{3}(R_{SM}^2 + R_{LAI}^2 + R_{PH}^2) \quad (29)$$

Terminal energy was evaluated as an uncalibrated retrieval-quality diagnostic. For each iEBT prediction, the final energy after refinement was compared with absolute prediction error. Diagnostic analyses included Spearman correlation between terminal energy and absolute error, RMSE after excluding the highest-energy samples, and AUROC for identifying high-error predictions. High-error samples were defined as the upper quartile of absolute residuals for each target.

## 3. Results

### 3.1. Random-split performance of learned retrieval models

The random-split benchmark evaluated four retrieval models using the same Sentinel-1, Sentinel-2, and temporal feature set: RF, TT, tEBT, and the proposed iEBT. Performance was assessed on the held-out test set using the $R^2$, RMSE. Table 4 reports the main quantitative benchmark as the mean performance across four random seeds for the neural models. Because Figure 4 represents a single trained realization, panel-level $R^2$ values may differ slightly from the multi-seed benchmark values reported in Table 4. Model ranking and statistical comparisons are based on the four-seed averages reported in Table 4.

Across the three target variables, iEBT achieved the highest mean performance across four random seeds, with average $R^2 = 0.854 \pm 0.012$. The tEBT model ranked second with an average $R^2 = 0.775 \pm 0.003$, followed by TT with average $R^2 = 0.749 \pm 0.006$. RF achieved an average $R^2 = 0.728$, indicating that the classical ensemble model remained competitive despite lacking explicit modality-specific encoding, energy-based training, or iterative refinement.

At the target level, LAI was the most accurately retrieved variable across the learned models. The iEBT model performed well with an $R^2$ of 0.905 for LAI, demonstrating that wheat development is sensitive to reflectance and vegetation indices of Sentinel-2 which are related to canopy density, chlorophyll absorption, and canopy closure. The soil moisture retrieval was weaker than LAI retrieval, but remained positive for all models with iEBT having $R^2$= 0.836. Plant height performed as intermediate with $R^2$ values ranging from 0.652 to 0.821 among models. This indicates that plant height was retrieved indirectly from canopy spectral response, SAR scattering behavior, and seasonal progression.

Table 4. Predictive performance of learned retrieval models on the held-out random test split across four random seeds.

| **Model** | **Average ($R^2$)** | **SM ($R^2$)** | **LAI ($R^2$)** | **PH ($R^2$)** |
|---|---|---|---|---|
| RF | 0.728 | 0.701 | 0.772 | 0.652 |
| TT | 0.749 ± 0.006 | 0.727 | 0.812 | 0.708 |
| tEBT | 0.775 ± 0.003 | 0.766 | 0.831 | 0.728 |
| iEBT | 0.854 ± 0.012 | 0.836 | 0.905 | 0.821 |

**Notes:** SM = soil moisture; LAI = leaf area index; PH = plant height. Average $R^2$ for neural models is reported as mean ± standard deviation across four seeds. RF is reported as a single trained model.

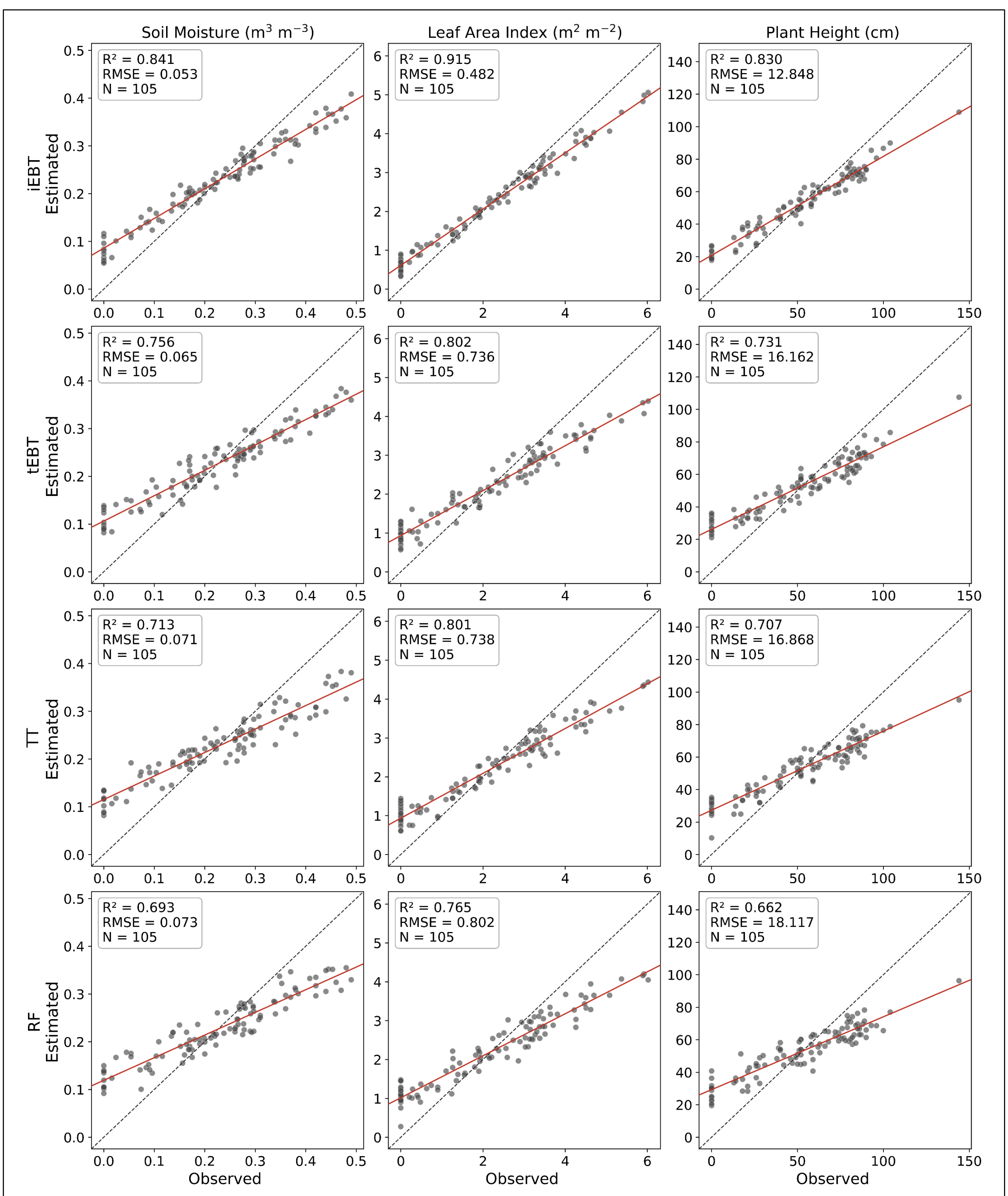

**Fig. 4** Observed versus estimated SM, LAI, and PH on the held-out test set for RF, TT, tEBT, and iEBT. The dashed line denotes the 1:1 relationship and the red line the fitted trend. Panel statistics are shown for a representative run; four-seed average performance is reported in Table 4.

Notes: Scatter plots show representative test-set predictions for soil moisture, LAI, and plant height

using iEBT, TT, tEBT, and RF. Columns correspond to target variables and rows correspond to retrieval models. The dashed black line represents the 1:1 relation between observed and estimated values, and the red line represents the fitted linear trend. Panel-level $R^2$, RMSE, and sample size are reported for the representative prediction set only. Table 4 provides the main four-seed benchmark used for quantitative model ranking.

Learned models showed that all the common gradients observed in the SM, LAI and PH were captured by representative scatter plots. The values for LAI were closely grouped around the 1:1 line, which is consistent with the high benchmark $R^2$ values. The estimation results were more diverse for SM, particularly for intermediate moisture levels, as the surface soil moisture, canopy attenuation, irrigation effects, and roughness- related responses in the C-band SAR are difficult to disentangle. PH estimates showed moderate scatter and sensitivity to outlying observations, which is expected because plant height is not directly measured by either Sentinel-1 or Sentinel-2.

### 3.2. WCM and PROSAIL baselines

The Water Cloud Model was used as a semi-empirical Sentinel-1 reference for evaluating SAR-only retrieval of LAI, SM, and PH. WCM parameters were calibrated independently for VV and VH polarizations using bounded least-squares optimization. The WCM evaluation used 700 cleaned field-satellite observations, including 490 training, 105 validation, and 105 test samples.

The WCM results were positive for all three target variables (Fig 5, Table 5). For LAI retrieval, VH performed better than VV, reaching $R^2 = 0.570$ and RMSE = 1.109 $m^2/m^2$, compared with $R^2 = 0.460$ and RMSE = 1.242 $m^2/m^2$ for VV. For SM retrieval, VV performed better than VH, reaching $R^2 = 0.580$ and RMSE = 0.084 $m^3/m^3$, compared with $R^2 = 0.470$ and RMSE = 0.094 $m^3/m^3$ for VH. For PH retrieval, VH again performed better than VV, reaching $R^2 = 0.550$ and RMSE = 22.077 cm, compared with $R^2 = 0.440$ and RMSE = 24.628 cm for VV.

**Table 5:** Performance of physical and semi-empirical reference models on the held-out test set.

| Model | Target | Polarization/source | $R^2$ | RMSE |
|---|---|---|---|---|
| WCM | LAI | VH | 0.570 | 1.109 $m^2/m^2$ |
| | | VV | 0.460 | 1.242 $m^2/m^2$ |
| | SM | VH | 0.470 | 0.094 $m^3/m^3$ |
| | | VV | 0.580 | 0.084 $m^3/m^3$ |
| | PH | VH | 0.550 | 22.077 cm |
| | | VV | 0.440 | 24.628 cm |
| PROSAIL/SNAP Biophysical Processor | LAI | Sentinel-2 | 0.519 | 0.912 $m^2/m^2$ |

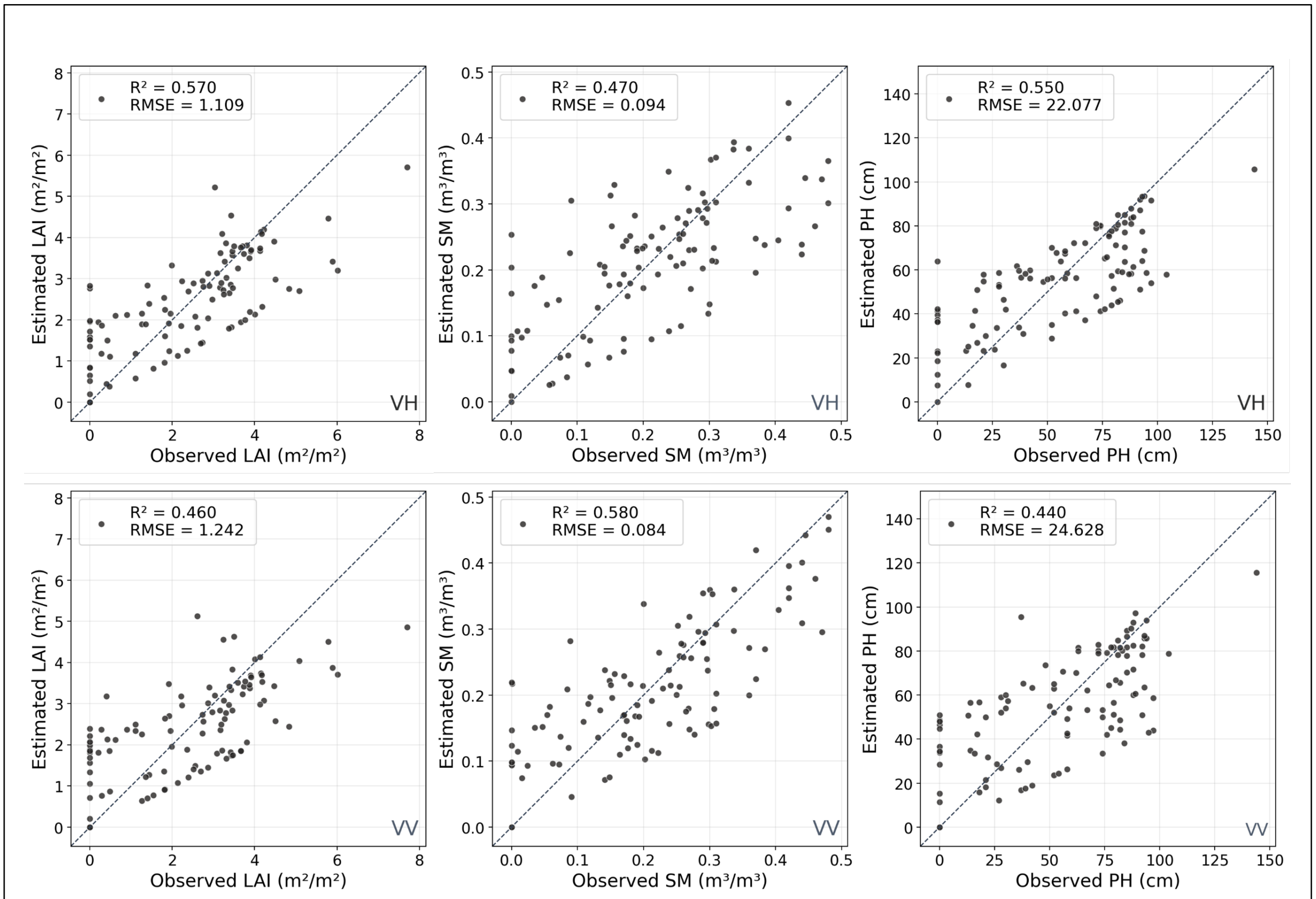

**Fig. 5.** Water Cloud Model observed-versus-estimated retrieval performance

Notes: The scatter plot shows WCM retrievals for LAI, soil moisture, and plant height using VH and VV polarizations on the WCM held-out test set. The top row shows VH-based retrievals and the bottom row shows VV-based retrievals. Columns correspond to LAI, SM, and PH. The dashed line represents the 1:1 relation between observed and estimated values. Panel labels report $R^2$ and RMSE.

The polarization-dependent WCM behavior is consistent with C-band scattering physics. VV retained stronger sensitivity to soil-surface and moisture-related scattering, which explains its stronger SM retrieval. VH showed stronger sensitivity to vegetation volume scattering and canopy structural development, which explains its stronger LAI and PH retrieval. These results indicate that WCM captures physically meaningful SAR responses, but SAR-only semi-empirical inversion remains less accurate than multimodal SAR-optical-temporal learning for joint wheat biophysical retrieval.

PROSAIL was retained as an optical physical reference for LAI. It was not treated as a multi-target competitor because the implemented PROSAIL/SNAP BiophysicalOp workflow does not retrieve SM or PH. Therefore, PROSAIL is used in the manuscript as an LAI-oriented optical reference, while WCM provides the SAR-oriented physical reference.

### 3.3. Spatial comparison of learned and physical retrievals

Spatial prediction maps were used to examine whether the learned and physical retrieval approaches produced coherent field-scale patterns. Figure 6 compares LAI, SM, and PH retrieval maps from iEBT, RF, WCM, and PROSAIL where relevant. The spatial maps complement the point-based validation by showing how each model behaves across the image domain.

The spatial maps provide a qualitative spatial consistency check that complements the point-based benchmark. The iEBT maps show coherent field-scale spatial patterns for LAI, SM, and PH, with low roughness and no evident saturation in the displayed panels. RF also preserves broad spatial gradients, but the maps show more local texture and roughness than iEBT, especially for SM and PH. WCM provides useful physical contrast but shows stronger artefacts in several panels. The WCM-VH LAI map shows high saturation, while the WCM-VV PH map is very smooth and has limited spatial differentiation. The PROSAIL LAI map provides an optical LAI reference, but it does not support joint interpretation of SM and PH.

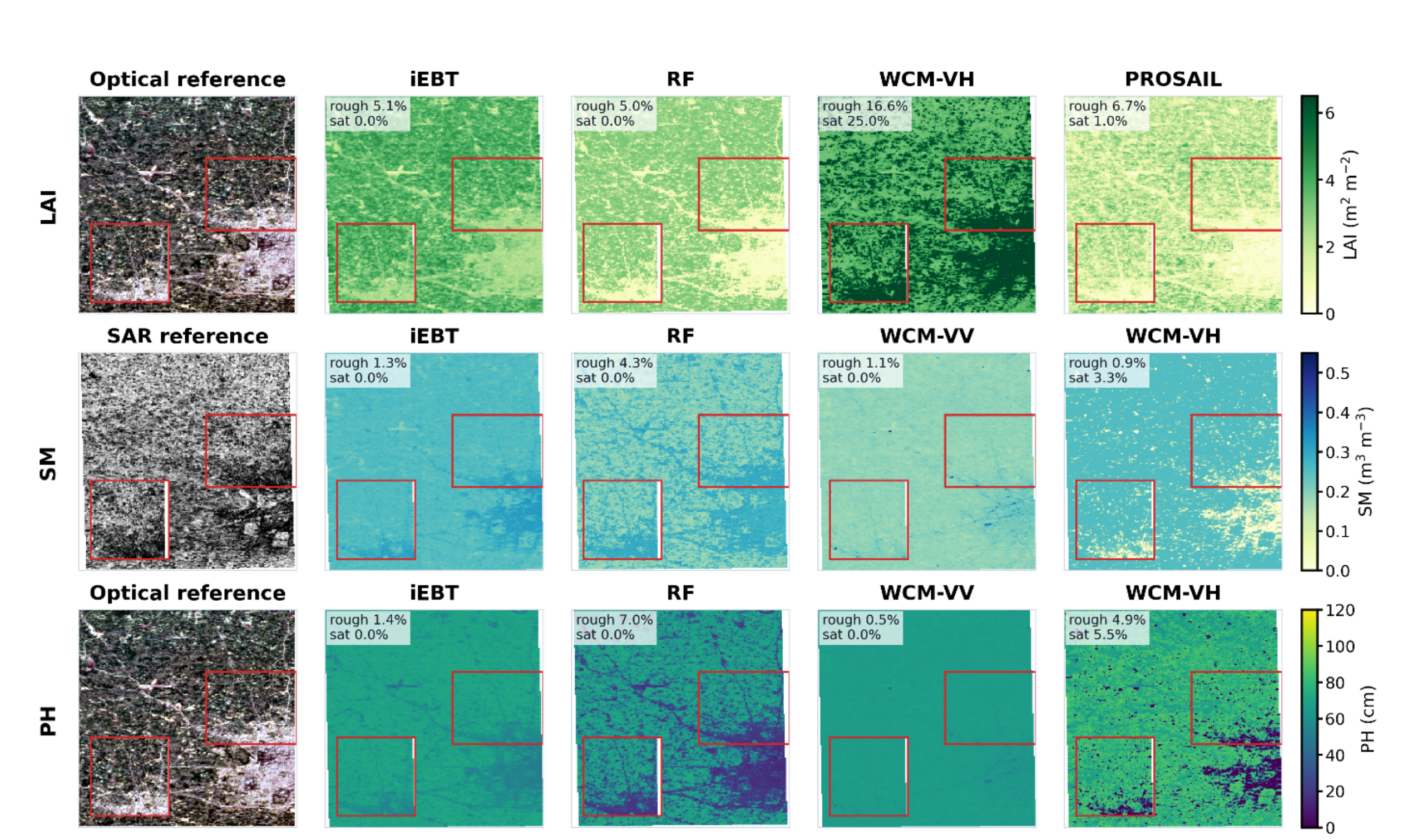

**Fig. 6** Spatial comparison of learned and physical retrieval maps

Notes: Rows correspond to LAI, soil moisture, and plant height. The first column shows the corresponding optical or SAR reference image used for visual context. Subsequent columns show retrieval maps from iEBT, RF, WCM-VV or WCM-VH, and PROSAIL where relevant. Red boxes highlight representative field areas used for visual inspection of roughness and saturation artefacts. Color scales are shared within each target variable to allow comparison across models.

These spatial results reinforce the quantitative findings. Physically motivated models are useful for interpretation and sensor-specific comparison, but the multimodal learned models provide more flexible spatial retrieval of the coupled [SM, LAI, PH] state. The iEBT maps are therefore interpreted as qualitative examples of field-scale retrieval behavior, while WCM and PROSAIL are retained as physical reference products.

### 3.4. Modality contribution and cross-campaign transferability

The modality ablation experiment evaluated whether SAR-only, optical-only, or full SAR-optical-temporal input produced the strongest retrieval performance as shown in Table 6A. Across the average of SM, LAI, and PH, full multimodal fusion produced the highest iEBT performance, reaching average $R^2 = 0.854 \pm 0.012$. SAR-only iEBT achieved average $R^2 = 0.786 \pm 0.007$, while optical-only iEBT achieved average $R^2 = 0.772 \pm 0.010$. These results show that Sentinel-1 and Sentinel-2 provide complementary information for joint crop-state retrieval.

The target-specific ablation behavior follows the expected sensor physics. SAR features contribute most strongly to SM retrieval because C-band backscatter is sensitive to soil dielectric properties and surface moisture conditions. Optical features dominate LAI retrieval because visible and near-infrared reflectance directly respond to canopy greenness, chlorophyll absorption, and canopy closure. PH depends on mixed information because plant height is not directly measured by either sensor; it must be inferred from vegetation volume scattering, optical canopy development, and season timing.

Leave-one-campaign-out validation was used to evaluate transferability across campaigns. The LOCO results are positive for all held-out campaigns, but performance varies substantially across years as shown in Table 6B. The 2019-2020 holdout was the most difficult, with iEBT reaching average $R^2 = 0.273 \pm 0.046$. The 2023 and 2024 holdouts showed stronger transfer performance, with iEBT reaching average $R^2 = 0.701 \pm 0.112$ and $R^2 = 0.725 \pm 0.064$, respectively. These results show that the model can transfer across campaigns, but the strength of transfer depends on the held-out year.

**Table 6.** Robustness experiments: modality ablation and leave-one-campaign-out validation.

**A.** Modality ablation: average $R^2$across SM, LAI, and PH

| Feature set | RF avg. ($R^2$) | TT avg. ($R^2$) | tEBT avg. ($R^2$) | iEBT avg. ($R^2$) |
|---|---|---|---|---|
| SAR only | 0.610 | 0.671 ± 0.015 | 0.727 ± 0.015 | 0.786 ± 0.007 |
| Optical only | 0.663 | 0.683± 0.018 | 0.730 ± 0.011 | 0.772 ± 0.010 |
| Full multimodal | 0.728 | 0.749 ± 0.016 | 0.775 ± 0.014 | 0.854 ± 0.012 |

**B.** Leave-one-campaign-out validation: average $R^2$

| Held-out campaign | Train n | Test n | RF | TT | tEBT | iEBT |
|---|---|---|---|---|---|---|
| 2019-2020 | 516 | 184 | 0.176 | 0.226 ± 0.080 | 0.192 ± 0.052 | 0.273 ± 0.046 |
| 2023 | 375 | 325 | 0.650 | 0.603 ± 0.043 | 0.658 ± 0.159 | 0.701 ± 0.112 |

| 2024 | 509 | 191 | 0.447 | 0.627 ± 0.061 | 0.662 ± 0.023 | 0.725 ± 0.064 |
|---|---|---|---|---|---|---|

**Notes:** Panel A reports average random-split $R^2$across the three target variables. Panel B reports average $R^2$for the held-out campaign.

The robustness experiments separate within-distribution retrieval from cross-campaign generalization. The random split shows that the models retrieve wheat biophysical variables effectively within the observed data distribution. The LOCO experiment shows that campaign-level transfer is more difficult, especially for the 2019-2020 holdout. This distinction is important for precision agriculture because operational retrieval models must generalize across seasons, irrigation regimes, phenological distributions, and field-management conditions.

### 3.5. Iterative energy refinement and retrieval diagnostics

The refinement analysis evaluated whether energy-based inference improved the initial crop-state proposal. For the representative iEBT run shown in Fig. 7, refinement increased SM $R^2$ from 0.721 to 0.796 and LAI $R^2$ from 0.792 to 0.915, while PH changed only slightly from $R^2 = 0.772$ to $R^2 = 0.768$. These results indicate that the refinement step mainly improved SM and LAI retrieval, whereas PH benefited less from energy-based updating.

The sample refinement also shows that the energy module does not replicate the proposal head. Rather the model makes a direct crop-state proposal, and further modifies the candidate [SM, LAI, PH] through energy minimization. Unlike a direct regressor, this retrieval mechanism is due to the evaluation of the compatibility of the proposed crop-state with the SAR-optical-temporal observation context.

The refinement trajectories indicate that with the energy gradient learned, the proposed crop states progress towards the expected target in the model's context. For SM and LAI, the proposed refinement brought the values into a 1:1 scatter relationship, while PH showed a smaller relative change. This indicates that for structural crop variables, more height-sensitive information may be required. The spatial error panels show that residuals were spread out across the campaigns rather than being concentrated in a specific sampling year, though in some locations, the target variable was estimated with either a persistent over or under estimation.

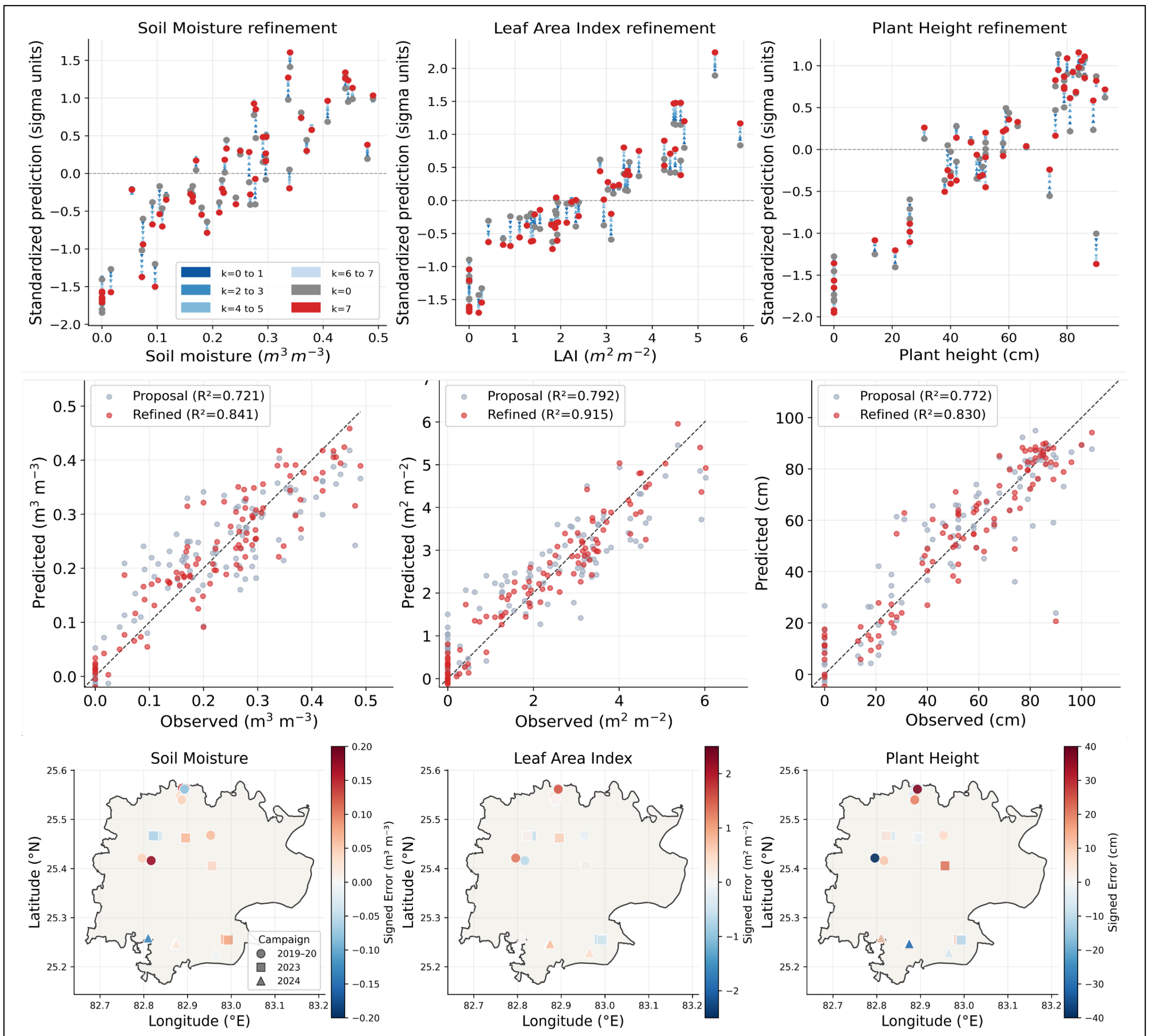

**Fig. 7.** Iterative energy-based refinement behavior for a representative iEBT run. The top row shows standardized candidate predictions for SM, LAI, and PH across refinement steps, with the initial proposal and subsequent updates shown separately. The middle row compares observed values with the initial proposal and final refined predictions; the dashed line is the 1:1 relation. The bottom row maps signed residuals for each target, with marker shape indicating campaign. Positive residuals indicate overestimation and negative residuals indicate underestimation.

Terminal energy was evaluated as an uncalibrated retrieval-compatibility diagnostic. Higher terminal energy indicates lower compatibility between the final predicted crop state and the multimodal observation context. Table 7 summarizes the diagnostic relationship between terminal energy and prediction residuals.

**Table 7. Energy diagnostic statistics**

| Diagnostic | $SM (m^3/m^3)$ | $LAI (m^2/m^2)$ | PH (cm) | Interpretation |
|---|---|---|---|---|
| Full-test RMSE baseline | 0.059 | 0.483 | 14.42 | Baseline RMSE before energy-based screening |
| Spearman $\rho$ : terminal energy vs. absolute error | 0.38 | 0.46 | 0.35 | Moderate positive association between energy and residual magnitude |
| Remove top 5% energy | 0.056 | 0.455 | 13.85 | Low-compatibility samples have higher residuals |
| Remove top 10% energy | 0.053 | 0.431 | 13.25 | Main post-retrieval screening result |
| Remove top 20% energy | 0.050 | 0.401 | 12.45 | Stronger screening improves accuracy with reduced coverage |
| AUROC for identifying high-error samples | 0.71 | 0.76 | 0.70 | Terminal energy provides fair high-error detection |

The diagnostic results show that terminal energy contains useful information about retrieval reliability. Spearman correlations between terminal energy and absolute residuals were positive for all three targets, with the strongest association for LAI. Removing the 10% highest-energy samples reduced RMSE from 0.059 to 0.053 $m^3\ m^{-3}$ for SM, from 0.483 to 0.431 $m^2\ m^{-2}$ for LAI, and from 14.42 to 13.25 cm for PH. The 20% screening level further reduced RMSE for all targets. AUROC values between 0.70 and 0.76 show that terminal energy can identify high-error samples better than random ranking. These results support the use of terminal energy as a practical post-retrieval quality-screening diagnostic.

## 4. Discussion

### 4.1. Compatibility-guided SAR-optical retrieval

This study shows that joint retrieval of wheat soil moisture, LAI, and plant height benefits from treating SAR-optical fusion as a compatibility-guided retrieval problem rather than only as direct regression. The proposed iEBT achieved the strongest numerical performance on the random test split, but its main contribution is the retrieval mechanism: an initial [SM, LAI, PH] proposal is refined by minimizing a learned energy function conditioned on Sentinel-1, Sentinel-2, and temporal descriptors (LeCun et al., 2006; Gladstone et al., 2025).

Care should be taken in making this distinction because a given satellite response can represent a variety of soil water status, canopy density, surface roughness, plant height and phenological stage. The iEBT test takes the form of a learned inversion test, compared to the conventional feed-forward prediction test of Belanger & McCallum (2016), because it checks whether a candidate crop state is compatible with the multimodal observations. This is substantiated by the proposal-to-refinement analysis which showed that generally SM and LAI were improved with refinement, whereas PH was not improved consistently. However, this is physically plausible as SM and LAI have more direct relationship to Sentinel-1 backscatter and Sentinel-2 reflectance, while PH is an indirect representation via canopy structure, volume scattering of vegetation and seasonal progression (Harfenmeister et al., 2019). The refinement-depth analysis indicates that K = 8 was an appropriate operating point for this dataset, balancing terminal energy reduction and retrieval accuracy.

### 4.2. Sensor information content and physical interpretation

The results of the experiments within the modality-ablation and physical-baseline experiments are consistent and show that Sentinel-1 and Sentinel-2 provide complementary but target-specific information (Ma et al., 2022). LAI was retrieved most accurately in agreement with its sensitivity to visible, red-edge and near-infrared reflectance, which reflect the greenness, the ability to absorb chlorophyll and the density of the vegetation canopy, respectively (Xie et al., 2018). Optical only retrieval consequently reached full fusion performance for LAI, suggesting that Sentinel-2 delivers the main canopy-density signal. For SM, however, it was more sensitive to SAR input, as anticipated, given the sensitivity of C-band backscatter to near surface moisture and soil dielectric properties (Yang et al., 2021). The acquisition geometry, surface roughness, irrigation time, vegetation water content, canopy attenuation, however, also affect the SAR backscatter, with the result that SM was more challenging than LAI (Wang et al., 2020).

PH was the most indirectly retrieved target because neither Sentinel-1 nor Sentinel-2 directly measures canopy height. Although VH backscatter and optical vegetation indices contain structural and phenological information, future PH improvement may require stronger structural descriptors such as multi-date SAR texture, incidence-angle-normalized backscatter, crop-stage constraints, or complementary height-sensitive observations. The WCM and PROSAIL comparisons help connect the learned results to remote-sensing physics. WCM showed meaningful polarization behavior, with VV more informative for SM and VH more useful for LAI and PH, but remained limited as a SAR-only semi-empirical inversion. PROSAIL provides an optical reference for LAI, but cannot retrieve SM or PH in this study (Tomíček et al., 2021). These results support the value of physical models for interpretation while showing why multimodal learned retrieval is needed for simultaneous estimation of SM, LAI, and PH.

### 4.3. Terminal energy and retrieval reliability

A useful feature of the energy-based formulation is that it produces a terminal energy value after refinement. The value is not a calibrated uncertainty nor is it intended to be used as a confidence interval, but rather as an uncalibrated compatibility score that reflects the degree to which the final [SM, LAI, PH] state is consistent with the observed SAR, optical and temporal context. The diagnostic analysis results indicated that the larger the terminal energy, the larger the residuals, and the elimination of higher energy samples led to an improvement of RMSE for all three targets. Terminal energy can be used after retrieval to flag predictions that should be interpreted cautiously before operational use.

The reason it is practically important is that residual cloud contamination, SAR speckle and roughness effects, irrigation events, temporal mismatch between the crop and the satellite observations and inter-seasonal domain shift affect this crop retrieval (Levitan et al., 2019). A point prediction does not suggest that the observation pattern is abnormal or inconsistent. There is a missing layer of diagnosis in terminal energy that allows the identification of predictions that should be treated with caution; but further validation is required before such energy estimates can be used as formal uncertainty estimates. Future work should explore calibration strategies such as conformal prediction, model ensembles, or physics-informed residual modeling (Kakhani et al., 2024).

### 4.4. Transferability, limitations, and implications

Results from the random-split experiment illustrate a good within-distribution retrieval, while the leave-one-campaign-out experiment shows a cross-season transfer remains difficult. This is important because while random splitting tests interpolation within the data distribution observed, the campaign holdout tests generalization across seasons, phenological trajectories and irrigation patterns, and across the target distribution. The variable LOCO performance suggests that deployment of the operation will need to be additionally calibrated and validated independently over a broader area of years.

There are a number of caveats to note. The dataset size is still small for deep multimodal learning, particularly for inter-annual generalization testing. Secondly, campaigns have different phenological coverage and target distributions, which impact random-split and campaign-holdout evaluation. Third, the predictor set is small, and does not contain potentially valuable predictors such as SAR texture, incidence-angle normalization, multi-temporal growth descriptors, or field-level spatial context. Despite these limitations, this study shows that joint retrieval of SM, LAI, and PH can provide a more complete description of wheat condition than single-variable monitoring. The most defensible conclusion is that iEBT offers a structured and interpretable retrieval framework with competitive accuracy and added reliability diagnostics instead of a a universally superior predictor in all validation settings.

## 5. Conclusion

This study developed an iterative energy-based multimodal transformer for joint retrieval of wheat surface soil moisture, leaf area index, and plant height from Sentinel-1 SAR, Sentinel-2 optical, and temporal observations. The proposed iEBT framework treats SAR-optical fusion as a compatibility-guided retrieval problem rather than a single-step regression task. By first generating an initial crop-state proposal and then refining the candidate [SM, LAI, PH] state through minimization of a learned energy function, the model provides a learned inversion mechanism that evaluates whether predicted crop states are consistent with the combined microwave, optical, and phenological evidence.

Using 700 quality-controlled field-satellite observations from wheat campaigns in Varanasi, India, iEBT achieved the highest numerical performance among the evaluated learned models on the random test split, reaching a four-seed average $R^2$ of 0.854 ± 0.012. The refinement analysis showed that energy-based inference improved the initial proposal mainly for soil moisture and LAI, while plant height was less consistently refined, reflecting the weaker direct sensitivity of Sentinel-1 and Sentinel-2 to crop height. The modality-ablation results confirmed the expected sensor physics: Sentinel-1 contributed most strongly to surface soil-moisture retrieval, Sentinel-2 dominated LAI estimation, and plant height depended on combined SAR structural, optical canopy, and temporal information.

The comparison with WCM and PROSAIL showed that physical and semi-empirical retrieval models remain valuable for interpretation, especially for understanding polarization-specific SAR responses and optical LAI retrieval. However, their single-sensor structure limits their ability to jointly retrieve the coupled soil-anopy state across heterogeneous wheat fields. In contrast, the proposed multimodal framework provides a more flexible route for simultaneous retrieval of SM, LAI, and PH. The terminal energy produced after refinement also provided a useful, uncalibrated diagnostic of retrieval quality: higher energy was associated with larger residuals, and screening high-energy predictions reduced RMSE across all three target variables. Nevertheless, leave-one-campaign-out validation showed that cross-season transfer remains a major challenge. The model performed well within the observed data distribution, but transferability varied across campaigns, emphasizing the need for larger multi-year datasets, independent site validation, and improved handling of phenological and management-driven domain shifts.

Overall, this study demonstrates that compatibility-guided SAR-optical fusion is a promising direction for field-scale wheat biophysical retrieval, particularly when prediction accuracy, physical interpretability, and post-retrieval reliability screening are all required for precision-agriculture applications.

**Declaration of AI-Assisted Writing**

During the preparation of this work, the authors used LLM for language editing and pre-submission manuscript auditing. After using this tool, the authors reviewed and edited the content as needed and take full responsibility for the content of the submitted manuscript.

**Declaration of competing interest**

The authors declare that they have no known competing financial interests or personal relationships that could have appeared to influence the work reported in this paper.

**Acknowledgements**

The authors gratefully acknowledge financial support from the Indian Space Research Organization (ISRO) through Project Grant No. R&D/SA/ISRO/PHY/22-23/03/392. The authors also acknowledge support from the Land Cover Land Use Change Program (LCLUC) of the National Aeronautics and Space Administration (NASA) under Grant No. 80NSSC20K0740. The authors thank the field teams and collaborating institutions involved in field data collection, satellite data processing, and logistical support for the wheat campaigns in Varanasi, India.

## References

Ahmad, S., Kumar, S. V., Lahmers, T. M., Wang, S., Liu, P., Wrzesien, M. L., Bindlish, R., Getirana, A., Locke, K., Holmes, T., & Otkin, J. A. (2022). Flash drought onset and development mechanisms captured with soil moisture and vegetation data assimilation. *Water Resources Research, 58*(12). https://doi.org/10.1029/2022wr032894

Alliès, A., Roumiguié, A., Dejoux, J.-F., Fieuzal, R., Jacquin, A., Veloso, A., Champolivier, L., & Baup, F. (2021). Evaluation of multiorbital SAR and multisensor optical data for empirical estimation of rapeseed biophysical parameters. *IEEE Journal of Selected Topics in Applied Earth Observations and Remote Sensing, 14*, 7268–7283. https://doi.org/10.1109/JSTARS.2021.3095537

Ayehu, G., Tadesse, T., Gessesse, B., Yigrem, Y., & M. Melesse, A. (2020). Combined use of sentinel-1 sar and landsat sensors products for residual soil moisture retrieval over agricultural fields in the upper Blue Nile basin, Ethiopia. *Sensors*, *20*(11), 3282.

Bahrami, H., McNairn, H., Mahdianpari, M., & Homayouni, S. (2022). A meta-analysis of remote sensing technologies and methodologies for crop characterization. *Remote Sensing, 14*(22), 5633. https://doi.org/10.3390/rs14225633

Bateni, S. M., Entekhabi, D., & Castelli, F. (2013). Mapping evaporation and estimation of surface control of evaporation using remotely sensed land surface temperature from a constellation of satellites. *Water Resources Research, 49*(2), 950–968. https://doi.org/10.1002/wrcr.20071

Belanger, D., & McCallum, A. (2016). Structured prediction energy networks. In *Proceedings of the 33rd International Conference on Machine Learning* (Vol. 48, pp. 983–992). PMLR.

Bouchat, J., Tronquo, E., Orban, A., Neyt, X., Verhoest, N. E. C., & Defourny, P. (2022). Green area index and soil moisture retrieval in maize fields using multi-polarized C- and L-band SAR data and the Water Cloud Model. *Remote Sensing, 14*(10), 2496. https://doi.org/10.3390/rs14102496

Bouras, E. H., Jarlan, L., Er-Raki, S., Albergel, C., Richard, B., Balaghi, R., & Khabba, S. (2020). Linkages between rainfed cereal production and agricultural drought through remote sensing indices and a land data assimilation system: A case study in Morocco. *Remote Sensing, 12*(24), 4018. https://doi.org/10.3390/rs12244018

Bousbih, S., Zribi, M., Lili-Chabaane, Z., Baghdadi, N., Hajj, M. E., Gao, Q., & Mougenot, B. (2017). Potential of Sentinel-1 radar data for the assessment of soil and cereal cover parameters. *Sensors, 17*(11), 2617. https://doi.org/10.3390/s17112617

Chauhan, S., Srivastava, H. S., & Patel, P. (2018). Wheat crop biophysical parameters retrieval using hybrid-polarized RISAT-1 SAR data. *Remote Sensing of Environment, 216*, 28–43. https://doi.org/10.1016/j.rse.2018.06.014

Corbari, C., Jovanovic, D. S., Nardella, L., Sobrino, J. A., & Mancini, M. (2020). Evapotranspiration estimates at high spatial and temporal resolutions from an energy-water balance model and satellite data in the Capitanata Irrigation Consortium. *Remote Sensing, 12*(24), 4083. https://doi.org/10.3390/rs12244083

Das, D. P., & Pandey, A. (2024). Soil moisture retrieval from dual-polarized Sentinel-1 SAR data over agricultural regions using a water cloud model. *Environmental Monitoring and Assessment*, *197*(1), 52. https://doi.org/10.1007/s10661-024-13510-4

Frappart, F., Wigneron, J., Li, X., Liu, X., Al-Yaari, A., Fan, L., Wang, M., Moisy, C., Masson, E. L., Lafkih, Z. A., Vallé, C., Ygorra, B., & Baghdadi, N. (2020). Global monitoring of the vegetation dynamics from the vegetation optical depth (VOD): A review. *Remote Sensing, 12*(18), 2915. https://doi.org/10.3390/rs12182915

Ghosh, S. S., Dey, S., Bhogapurapu, N., Homayouni, S., Bhattacharya, A., & McNairn, H. (2022). Gaussian process regression model for crop biophysical parameter retrieval from multi-polarized C-band SAR data. *Remote Sensing, 14*(4), 934. https://doi.org/10.3390/rs14040934

Gladstone, A., Nanduru, G., Islam, M. M., Han, P., Ha, H., Chadha, A., ... & Iqbal, T. (2025). Energy-based transformers are scalable learners and thinkers. *arXiv preprint arXiv:2507.02092*. https://doi.org/10.48550/arXiv.2507.02092

Gou, Y., Ryan, C. M., & Reiche, J. (2022). Large area aboveground biomass and carbon stock mapping in woodlands in Mozambique with L-band radar: Improving accuracy by accounting for soil moisture effects using the Water Cloud Model. *Remote Sensing, 14*(2), 404. https://doi.org/10.3390/rs14020404

Hajj, M. E., Baghdadi, N., Zribi, M., Belaud, G., Cheviron, B., Courault, D., & Ruget, F. (2016). Soil moisture retrieval over irrigated grassland using X-band SAR data. *Remote Sensing of Environment, 176*, 202–218. https://doi.org/10.1016/j.rse.2016.01.027

Han, L., Wang, C., Yu, T., Gu, X., & Liu, Q. (2020). High-precision soil moisture mapping based on multi-model coupling and background knowledge, over vegetated areas using Chinese GF-3 and GF-1 satellite data. *Remote Sensing, 12*(13), 2123. https://doi.org/10.3390/rs12132123

Harfenmeister, K., Spengler, D., & Weltzien, C. (2019). Analyzing temporal and spatial characteristics of crop parameters using Sentinel-1 backscatter data. *Remote Sensing, 11*(13), 1569. https://doi.org/10.3390/rs11131569

Hosseini, M., McNairn, H., Mitchell, S., Robertson, L. D., Davidson, A., Ahmadian, N., Bhattacharya, A., Borg, E., Conrad, C., Dąbrowska-Zielińska, K., Abelleyra, D. de, Gurdak, R., Kumar, V., Kussul, N., Mandal, D., Rao, Y. S., Saliendra, N. Z., Shelestov, A., Spengler, D., … Becker-Reshef, I. (2021). A comparison between support vector machine and Water Cloud Model for estimating crop leaf area index. *Remote Sensing, 13*(7), 1348. https://doi.org/10.3390/rs13071348

Kakhani, N., Alamdar, S., & Kebonye, N. M. (2024). Uncertainty quantification of soil organic carbon estimation from remote sensing data with conformal prediction. *Remote Sensing, 16*(3), 438. https://doi.org/10.3390/rs16030438

LeCun, Y., Chopra, S., Hadsell, R., Ranzato, M., & Huang, F. J. (2006). A tutorial on energy-based learning. In G. Bakir, T. Hofmann, B. Schölkopf, A. J. Smola, B. Taskar, & S. V. N. Vishwanathan (Eds.), *Predicting structured data* (pp. 191–246). MIT Press.

Levitan, N., Kang, Y., & Özdoğan, M. (2019). Evaluation of the uncertainty in satellite-based crop state variable retrievals due to site and growth stage specific factors and their potential in coupling with crop growth models. *Remote Sensing, 11*(16), 1928. https://doi.org/10.3390/rs11161928

Li, J., & Wang, S. (2018). Using SAR-derived vegetation descriptors in a Water Cloud Model to improve soil moisture retrieval. *Remote Sensing, 10*(9), 1370. https://doi.org/10.3390/rs10091370

Lin, J., Shen, Q., Wu, J., Zhao, W., & Liu, L. (2022). Assessing the potential of downscaled far red solar-induced chlorophyll fluorescence from the canopy to leaf level for drought monitoring in winter wheat. *Remote Sensing, 14*(6), 1357. https://doi.org/10.3390/rs14061357

Liu, C., Chen, Z., Shao, Y., Chen, J., Tuya, H., & Pan, H. (2019). Research advances of SAR remote sensing for agriculture applications: A review. *Journal of Integrative Agriculture, 18*(3), 506–525. https://doi.org/10.1016/S2095-3119(18)62016-7

Liu, J., Xu, Y., Li, H., & Guo, J. (2021). Soil moisture retrieval in farmland areas with Sentinel multi-source data based on regression convolutional neural networks. *Sensors, 21*(3), 877. https://doi.org/10.3390/s21030877

Liu, X., Liu, X., Li, X., Zhang, X., Nian, L., Zhang, X., Wang, P., Ma, B., Li, Q., Zhang, X., Hui, C., Bai, Y., Jin, B., Zhang,X., Liu, J., Sun, J., Yu, W., & Luo, L. (2023). Retrieving soil moisture in the first-level tributary of the Yellow River–Wanchuan River Basin based on CD algorithm and Sentinel-1/2 data. *Water, 15*(19), 3409. https://doi.org/10.3390/w15193409

Ma, C., Johansen, K., & McCabe, M. F. (2022). Monitoring irrigation events and crop dynamics using Sentinel-1 and Sentinel-2 time series. *Remote Sensing, 14*(5), 1205. https://doi.org/10.3390/rs14051205

Meng, Q., Zhang, L., Xie, Q., Yao, S., Xu, C., & Zhang, Y. (2018). Combined use of GF-3 and Landsat-8 satellite data for soil moisture retrieval over agricultural areas using artificial neural network. *Advances in Meteorology, 2018*, 1–11. https://doi.org/10.1155/2018/9315132

Nduku, L., Munghemezulu, C., Mashaba-Munghemezulu, Z., Ratshiedana, P. E., Sibanda, S., & Chirima, G. (2024). Synergetic use of Sentinel-1 and Sentinel-2 data for wheat-crop height monitoring using machine learning. *Agriengineering, 6*(2), 1093–1116. https://doi.org/10.3390/agriengineering6020063

Pan, H., Chen, Z., Wit, A. de, & Ren, J. (2019). Joint assimilation of leaf area index and soil moisture from Sentinel-1 and Sentinel-2 data into the WOFOST model for winter wheat yield estimation. *Sensors, 19*(14), 3161. https://doi.org/10.3390/s19143161

Papadavid, G., & Toulios, L. (2017). The use of earth observation methods for estimating regional crop evapotranspiration and yield for water footprint accounting. *The Journal of Agricultural Science, 156*(5), 599–617. https://doi.org/10.1017/S0021859617000594

Park, S.-E., Jung, Y. T., Cho, J., Moon, H., & Han, S.-H. (2019). Theoretical evaluation of Water Cloud Model vegetation parameters. *Remote Sensing, 11*(8), 894. https://doi.org/10.3390/rs11080894

Pierdicca, N., Comite, D., Camps, A., Carreño-Luengo, H., Cenci, L., Clarizia, M. P., Costantini, F., Dente, L., Guerriero, L., Mollfulleda, A., Paloscia, S., Park, H., Santi, E., Zribi, M., & Floury, N. (2022). The potential of spaceborne GNSS reflectometry for soil moisture, biomass, and freeze–thaw monitoring: Summary of a European Space Agency-funded study. *IEEE Geoscience and Remote Sensing Magazine, 10*(2), 8–38. https://doi.org/10.1109/MGRS.2021.3115448

Santi, E., & Paloscia, S. (2019). *Microwave indices from active and passive sensors for remote sensing applications*. https://doi.org/10.3390/books978-3-03897-821-3

Singh, S. K., Prasad, R., Srivastava, P. K., Yadav, S. A., Yadav, V. P., & Sharma, J. (2023). Incorporation of first-order backscattered power in Water Cloud Model for improving the leaf area index and soil moisture retrieval using dual-polarized Sentinel-1 SAR data. *Remote Sensing of Environment, 296*, 113756.

Singh, S. K., Prasad, R., Yadav, S. A., Srivastava, P. K., Singh, G., & Srivastava, H. S. (2024). Fusion of optical and SAR data using three approaches for the estimation of LAI with modified integral equation model. *IEEE Geoscience and Remote Sensing Letters, 21*, 1–5.

Steele-Dunne, S., McNairn, H., Monsiváis-Huertero, A., Judge, J., Liu, P., & Papathanassiou, K. (2017). Radar remote sensing of agricultural canopies: A review. *IEEE Journal of Selected Topics* *in Applied Earth Observations and Remote Sensing, 10*(5), 2249–2273. https://doi.org/10.1109/JSTARS.2016.2639043

Tao, L., Wang, G., Chen, X., Li, J., & Cai, Q. (2019). Estimation of soil moisture using a vegetation scattering model in wheat fields. *Journal of Applied Remote Sensing, 13*(04), 1. https://doi.org/10.1117/1.JRS.13.4.044503

Tomíček, J., Misurec, J., Lukeš, P., Hanuš, J., & Homolová, L. (2021). Prototyping a generic algorithm for crop parameter retrieval across the season using radiative transfer model inversion and Sentinel-2 satellite observations. *Remote Sensing, 13*(18), 3659. https://doi.org/10.3390/rs13183659

Tu, L., & Gimpel, K. (2019). Benchmarking approximate inference methods for neural structured prediction. In *Proceedings of the 2019 Conference of the North American Chapter of the Association for Computational Linguistics: Human Language Technologies* (pp. 3313–3328). Association for Computational Linguistics. https://doi.org/10.18653/v1/N19-1335

Veloso, A., Mermoz, S., Bouvet, A., Toan, T. L., Planells, M., Dejoux, J.-F., & Ceschia, É. (2017). Understanding the temporal behavior of crops using Sentinel-1 and Sentinel-2-like data for agricultural applications. *Remote Sensing of Environment, 199*, 415–426. https://doi.org/10.1016/j.rse.2017.07.015

Vermunt, P., Steele-Dunne, S., Khabbazan, S., Kumar, V., & Judge, J. (2022). Towards understanding the influence of vertical water distribution on radar backscatter from vegetation using a multi-layer Water Cloud Model. *Remote Sensing, 14*(16), 3867. https://doi.org/10.3390/rs14163867

Wang, J., Li, P., Bi, R., Xu, L., He, P., Zhao, Y., & Li, X. (2024). Applicability of different assimilation algorithms in crop growth model simulation of evapotranspiration. *Agronomy, 14*(11), 2674. https://doi.org/10.3390/agronomy14112674

Wang, Q., Li, J., Jin, T., Chang, X., Zhu, Y., & Li, Y. (2020). Comparative analysis of Landsat-8, Sentinel-2, and GF-1 data for retrieving soil moisture over wheat farmlands. *Remote Sensing, 12*(17), 2708. https://doi.org/10.3390/rs12172708

Xie, Q., Dash, J., Huang, W., Peng, D., Qin, Q., Mortimer, H., Casa, R., Pignatti, S., Laneve, G., Pascucci, S., Dong, Y., & Ye, H. (2018). Vegetation indices combining the red and red-edge spectral information for leaf area index retrieval. *IEEE Journal of Selected Topics in Applied Earth Observations and Remote Sensing, 11*(5), 1482–1493. https://doi.org/10.1109/JSTARS.2018.2813281

Yang, M., Wang, H., Tong, C., Zhu, L., Deng, X., Deng, J., & Wang, K. (2021). Soil moisture retrievals using multi-temporal Sentinel-1 data over Nagqu region of Tibetan Plateau. *Remote Sensing, 13*(10), 1913. https://doi.org/10.3390/rs13101913

Zhang, L., Meng, Q., Yao, S., Wang, Q., Zeng, J., Zhao, S., & Ma, J. (2018). Soil moisture retrieval from the Chinese GF-3 satellite and optical data over agricultural fields. *Sensors, 18*(8), 2675. https://doi.org/10.3390/s18082675

Zhang, L., Meng, Q., Zeng, J., Wei, X., & Shi, H. (2021). Evaluation of Gaofen-3 C-band SAR for soil moisture retrieval using different polarimetric decomposition models. *IEEE Journal of Selected Topics in Applied Earth Observations and Remote Sensing, 14*, 5707–5719. https://doi.org/10.1109/JSTARS.2021.3083287

Zhuo, W., Huang, J., Li, L., Zhang, X., Ma, H., Gao, X., Huang, H., Xu, B., & Xiao, X. (2019). Assimilating soil moisture retrieved from Sentinel-1 and Sentinel-2 data into WOFOST model to improve winter wheat yield estimation. *Remote Sensing, 11*(13), 1618. https://doi.org/10.3390/rs11131618

Zribi, M., Sekhar, M., Bousbih, S., Bitar, A. A., Tomer, S. K., Baghdadi, N., & Bandyopadhyay, S. (2019). Analysis of L-band SAR data for soil moisture estimations over agricultural areas in the tropics. *Remote Sensing, 11*(9), 1122. https://doi.org/10.3390/rs11091122